\documentclass[letterpaper, 10 pt, conference]{ieeeconf}  

\IEEEoverridecommandlockouts                              

\usepackage{amsmath} 
\usepackage{amssymb}  
\usepackage{bm}
\usepackage{cite}
\newcommand{\vect}      [1] {\mathbf{#1}}
\newcommand{\mat}       [1] {\mathbf{#1}}
\newcommand{\equationdot}       {\ensuremath{\;.}}
\newcommand{\equationcomma}       {\ensuremath{\;,}}

\newcommand{\rsec}                                      [1]     {Section~\ref{#1}}
\newcommand{\reqn}                                      [1]     {(\ref{#1})}

\title{\LARGE \bf
Towards Continuous Time Finite Horizon LQR Control in SE(3)
}

\author{Shivesh Kumar$^{1}$, Andreas Mueller$^{2}$, Patrick Wensing$^{3}$ and Frank Kirchner$^{1,4}$
\thanks{This work was supported by RIMA
project (FKZ H2020-DT-2018-2020/H2020-DT-2018-1 \#824990) funded by
European Unions Horizon 2020 research and innovation programme. The
second author acknowledges the support of the LCM-K2 Center within the
framework of the Austrian COMET-K2 program.}
\thanks{$^{1}$Robotics Innovation Center, German Research Center for Artificial Intelligence (DFKI GmbH), Bremen, Germany.
        {\tt\small shivesh.kumar@dfki.de}}%
\thanks{$^{2}$Institute of Robotics, Johannes Kepler University, Linz, Austria
        {\tt\small a.mueller@jku.at}}%
\thanks{$^{3}$Department of Aerospace and Mechanical Engineering, University of Notre Dame, USA
        {\tt\small pwensing@nd.edu}}%
\thanks{$^{4}$Faculty of Mathematics and Computer Science, University of Bremen, Germany
        {\tt\small frank.kirchner@dfki.de}}%
}

\begin{document}

\maketitle
\thispagestyle{empty}
\pagestyle{empty}

\begin{abstract}

The control of free-floating robots requires dealing with several challenges. The motion of such robots evolves on a continuous manifold described by the Special Euclidean Group of dimension 3, known as $SE(3)$. Methods from finite horizon Linear Quadratic Regulators (LQR) control have gained recent traction in the robotics community. However, such approaches are inherently solving an unconstrained optimization problem and hence are unable to respect the manifold constraints imposed by the group structure of $SE(3)$. This may lead to small errors, singularity  problems and double cover issues depending on the choice of coordinates to model the floating base motion. In this paper, we propose the use of canonical exponential coordinates of $SE(3)$ and the associated Exponential map along with its differentials to embed this structure in the theory of finite horizon LQR controllers. 
\end{abstract}

\section{Introduction}
Methods from Lie Group and Screw theory are becoming increasingly popular in the robotics community to describe rigid body kinematics and dynamics~\cite{lynch2017modern, murray1994mathematical}. They are equally popular in the domain of robot control~\cite{bullo2005geometric}. Additionally, the geometric framework has been used in time integration schemes for general multi-body systems (MBS)~\cite{park2005geometric} including the flexible MBS~\cite{2015_valentin_thesis}.  

MBS motions evolve on a Lie group and their dynamics is naturally described by differential equations on that Lie group.
The most crucial Lie group for studying rigid body motion is the Special Euclidean Group of dimension 3, known as $SE(3)$. Its importance is shown by the fact that all possible rigid body motions are captured by subgroups of $SE(3)$. For the purpose of robot control or time integration of the robot dynamics, it is quite common to consider a direct product of the translational group $\mathbb{R}^3$ and the special orthogonal group $SO(3)$ i.e. $SO(3) \times \mathbb{R}^3$. This allows one to use different parameterizations for rotation (e.g., Euler angles, quaternions, etc.) and translation parts.  This configuration space (c-space) does, however, not account for the intrinsic geometry of rigid body motions in that rotations and translations are decoupled. However, rigid body motions in $SE(3)$ are mathematically defined as the semi-direct product between $SO(3)$ and $\mathbb{R}^3$ i.e. $SE(3) = SO(3) \ltimes \mathbb{R}^3$. In other words, $SO(3)$ is its quotient or factor group and $\mathbb{R}^3$ is its normal subgroup. In~\cite{MULLER20143}, it was shown that the naive use of $SO(3) \times \mathbb{R}^3$ can lead to constraint violations during geometric integration, which must be corrected by additional constraint stabilization techniques. By contrast, using $SE(3) = SO(3) \ltimes \mathbb{R}^3$ can often avoid such issues. Note that both $SE(3)$ and $SO(3) \times \mathbb{R}^3$ allow for the representation of rigid body configurations. But only $SE(3)$ allows for representing rigid body motions~\cite{MULLER20143}. 

Similarly, improper treatment of $SE(3)$ leads to various challenges in robot control. For example, the use of Euler angles to describe the rotation matrix leads to singularity issues~\cite{hemingway2018perspectives, shuster1993survey}. The use of quaternions addresses the singularity issue but the solutions may suffer from double cover issue i.e. multiple quaternion solutions to represent the same rotation matrix~\cite{diebel2006representing}. 
Geometric control~\cite{bullo2005geometric} attempts to unify the study of mechanics and control under the setting of differential
geometry. In~\cite{2015_koushi_sreenath}, a time varying LQR controller was developed using a variation-based linearization of the non-linear systems evolving on various Lie groups such as $SO(3)$ and $\mathbb{S}^2$. 
In~\cite{2021_planning_attitude}, the framework of geometric control was adopted to plan trajectories in $SO(3) \times \mathbb{R}^3$ with large attitude changes by exploiting the Cayley map of $SO(3)$ which involves the use of its local (non-canonical) coordinates. An extensive treatment of discrete-time differential dynamic programming (DDP) on Lie groups has been recently presented in~\cite{2021_Boutselis_discrete_DDP}. 
Another recent work~\cite{alcan2023trajectory} presents an approach for trajectory optimization on generic matrix Lie groups using an augmented Lagrangian-based constrained discrete Differential Dynamic Programming.
Finally,~\cite{teng2022lie} proposes the use of cost function design on the Lie algebra for control on Lie Groups. 

\paragraph*{Contribution} This paper proposes the use of canonical coordinates (screw coordinates) of $SE(3)$ to synthesize a finite horizon LQR controller for trajectory stabilization. By exploiting the exponential map and its differentials, we derive the linearization of the equation of motion (EOM) of a free-floating rigid body. The proposed linearization is easy to implement as it avoids the use of tensors by exploiting the directional derivative of the derivative of the exponential map (dexp map). Based on this linearization, a time varying LQR controller is developed. Note that we are currently evaluating the proposed controller in some numerical experiments and the results will be presented in the near future.

\paragraph*{Organization}
\rsec{sec_canonical_coordinates_SE3} presents the canonical screw coordinates of $SE(3)$ along with the exponential map and its differentials. \rsec{sec_geom_lin_eom} presents the equation of motion of a free-floating rigid body and its geometric linearization. \rsec{sec_tvlqr_se3} derives the corresponding finite horizon LQR controller. \rsec{sec_conc_outlook} concludes the paper and highlights our future work. 

\section{Canonical Coordinates on $SE(3)$, exp map and its Differentials}
\label{sec_canonical_coordinates_SE3}
This section presents the fundamentals for describing rigid body motion in $SE(3)$ in terms of canonical screw coordinates via the exponential map $\text{exp}$ and its differential $\text{dexp}$. Further, the directional derivative of the $\text{dexp}$ mapping is presented which is required for linearizing the EOM. For a detailed treatment, refer to~\cite{muller2021review}.

\subsection{Preliminaries}
Let $G$ be a $n$-dimensional Lie group with Lie algebra $\mathfrak{g}$. Lie
algebra elements are denoted $\hat{\mathbf{X}}\in \mathfrak{g}$, and when
represented as vectors, they are denoted $\mathbf{X}\in {\mathbb{R}}^{n}$,
which implies an obvious isomorphism. 
Let $\exp :\mathfrak{g}\rightarrow G$ be the exponential map on $G$. Its
right-trivialized differential $\mathrm{dexp}_{\hat{\mathbf{X}}}:\mathfrak{g}%
\rightarrow \mathfrak{g}$ is defined as%
\begin{equation}
\left( \mathrm{D}_{\hat{\mathbf{X}}}\exp \right) 
\hspace{-0.5ex}%
(\hat{\mathbf{Y}})=\mathrm{dexp}_{\hat{\mathbf{X}}}(\hat{\mathbf{Y}})\exp (%
\hat{\mathbf{X}})  \label{diff1}
\end{equation}%
where $\left( \mathrm{D}_{\hat{\mathbf{X}}}\exp \right) 
\hspace{-0.5ex}%
(\hat{\mathbf{Y}}):=\frac{d}{dt}\exp (\hat{\mathbf{X}}+t\hat{\mathbf{Y}}%
)|_{t=0}$ is the directional derivative $\mathrm{D}_{\hat{\mathbf{X}}}\exp :%
\mathfrak{g}\rightarrow T_{\exp \hat{\mathbf{X}}}G$ of $\exp $ at $\hat{%
\mathbf{X}}$ in direction of $\hat{\mathbf{Y}}$. The differential and its
inverse admit the series expansions \cite[pp. 26 \& 36ff]{hausdorff1906symbolische},%
\cite{iserles1984solving},\cite[Theorem 2.14.3.]{varadarajan2013lie} 
\begin{eqnarray}
\mathrm{dexp}_{\hat{\mathbf{X}}}(\hat{\mathbf{Y}}) &=&\sum_{i=0}^{\infty
}\frac{1}{\left( i+1\right) !}\mathrm{ad}_{\hat{\mathbf{X}}}^{i}(\hat{%
\mathbf{Y}})  \label{dexpSeries} \\
\mathrm{dexp}_{\hat{\mathbf{X}}}^{-1}(\hat{\mathbf{Y}})
&=&\sum_{i=0}^{\infty }\frac{B_{i}}{i!}\mathrm{ad}_{\hat{\mathbf{X}}}^{i}(%
\hat{\mathbf{Y}})  \label{dexpInvSeries}
\end{eqnarray}%
where $B_{i}$ denote the Bernoulli numbers.
In vector representation of $\mathfrak{g}$, the differential mapping attains
the form $\mathbf{dexp}_{\mathbf{X}}\mathbf{Y}$, with matrix $\mathbf{dexp}_{%
\mathbf{X}}$. This matrix and its inverse admit the series expansions%
\begin{eqnarray}
\mathbf{dexp}_{\hat{\mathbf{X}}} &=&\sum_{i=0}^{\infty }\frac{1}{\left(
i+1\right) !}\mathbf{ad}_{\hat{\mathbf{X}}}^{i}  \label{dexpMat} \\
\mathbf{dexp}_{\hat{\mathbf{X}}}^{-1} &=&\sum_{i=0}^{\infty }\frac{B_{i}}{%
i!}\mathbf{ad}_{\hat{\mathbf{X}}}^{i}.  \label{dexpInvMat}
\end{eqnarray}%
with the little adjoint matrix $\mathbf{ad}_{\mathbf{X}}$ such that $\mathrm{ad}_{%
\hat{\mathbf{X}}}(\hat{\mathbf{Y}})=\mathbf{\widehat{\mathbf{ad}_{%
\mathbf{X}}\mathbf{Y}}}$. 

\subsection{Rigid Body Motion in Terms of Exponential Map}
Pose of the rigid body is expressed in terms of the canonical coordinates $%
\mathbf{X}\in {\mathbb{R}}^{6}$ of the first kind with the exponential map $\mathbf{C}\left(
t\right) =\mathbf{C}_{0}\exp \hat{\mathbf{X}}\left( t\right) $, where $\hat{%
\mathbf{X}}\in se\left( 3\right) $. The closed form can be expressed for $%
\mathbf{X}=\left( \mathbf{x},\mathbf{y}\right) $ as%
\begin{align}
\exp (\hat{\mathbf{X}})& =\left( 
\begin{array}{cc}
\mathbf{R} & \ \ \tfrac{1}{\left\Vert \mathbf{x}\right\Vert ^{2}}(\mathbf{I}-%
\mathbf{R})\tilde{\mathbf{x}}\mathbf{y}+h\mathbf{y} \\ 
\mathbf{0} & 1%
\end{array}%
\right) ,\ \mathrm{for\ }\mathbf{x}\neq \mathbf{0}\   \notag \\
& =\left( 
\begin{array}{cc}
\mathbf{I} & \ \mathbf{y} \\ 
\mathbf{0} & 1%
\end{array}%
\right) ,\ \mathrm{for\ }\mathbf{x}=\mathbf{0}
\end{align}%
where%
\begin{equation}
\mat R = \exp \tilde{\mathbf{x}}=\mathbf{I}+\alpha \tilde{\mathbf{x}}+\tfrac{1}{2}%
\beta \tilde{\mathbf{x}}^{2}  \label{SO3exp6}
\end{equation}%
with $\alpha :=\mathrm{sinc}\left\Vert \mathbf{x}\right\Vert ,\ \ \beta :=%
\mathrm{sinc}^{2}\mathrm{\,}\frac{\left\Vert \mathbf{x}\right\Vert }{2}$.
The body twist in body-fixed representation is given in terms of the time
derivative of $\mathbf{X}$ by the \emph{local reconstruction equation}:
\begin{equation}
\mathbf{V}=\mathbf{dexp}_{-\mathbf{X}}\dot{\mathbf{X}}  
\label{RecExp}
\end{equation}%
where $\mathbf{dexp}_{\mathbf{X}}: \mathbb{R}^6 \rightarrow \mathbb{R}^6 $ is the matrix form of the right-trivialized differential of the exp map. The inverse
relation is $\dot{\mathbf{X}}=\mathbf{dexp}_{-\mathbf{X}}^{-1}\mathbf{V%
}$. For $SE(3)$, it can be expressed in closed-form as%
\begin{equation}
\mathbf{dexp}_{\mathbf{X}}^{-1}=\left( 
\begin{array}{cc}
\mathbf{dexp}_{\mathbf{x}}^{-1} & \ \ \mathbf{0} \\ 
(\mathrm{D}_{\mathbf{x}}\mathbf{dexp}^{-1})%
\hspace{-0.6ex}%
\left( \mathbf{y}\right)  & \mathbf{dexp}_{\mathbf{x}}^{-1}%
\end{array}%
\right)   \label{DexpInvSE3}
\end{equation}%
with 
\begin{align*}
(\mathrm{D}_{\mathbf{x}}\mathbf{dexp}^{-1})%
\hspace{-0.5ex}%
\left( \mathbf{y}\right) =-\tfrac{1}{2}\tilde{\mathbf{y}}+\frac{1}{%
\left\Vert \mathbf{x}\right\Vert ^{2}}\left( 1-\gamma \right) \left( \tilde{%
\mathbf{x}}\tilde{\mathbf{y}}+\tilde{\mathbf{y}}\tilde{\mathbf{x}}\right) +\\%
\tfrac{\mathbf{x}^{T}\mathbf{y}}{\left\Vert \mathbf{x}\right\Vert ^{4}}%
\left( \tfrac{1}{\beta }+\gamma -2\right) \tilde{\mathbf{x}}^{2}
\label{diffDexpInvSO3}
\end{align*}%
and $\gamma :=\alpha /\beta $~\cite{muller2021review}.
In vector representation $\vect X = (\vect x, \vect y) \in \mathbb{R}^6$, the matrix form of the adjoint operator for $SE(3)$ is 
\begin{equation}
 \mathbf{ad}_{\vect X} = 
\left( 
\begin{array}{cc}
\tilde{\mathbf{x}} &  \mathbf{0} \\ 
\tilde{\mathbf{y}} & \tilde{\mathbf{x}}%
\end{array}%
\right) \equationdot
\end{equation}
Using it, a computationally efficient and numerically stable version of \reqn{DexpInvSE3} was presented in~\cite{BOTTASSO1998307, bullo1995proportional} and is given by
\begin{equation}
\begin{aligned}
\mathbf{dexp}_{\mathbf{X}}^{-1}& =\mathbf{I}-\frac{1}{2}\mathbf{ad}_{\mathbf{X}}+ 
\tfrac{1}{\left\Vert 
\mathbf{x}\right\Vert ^{2}}\left( 2-\tfrac{1+3\alpha }{2\beta }\right) 
\mathbf{ad}_{\mathbf{X}}^{2}+ \\
& \tfrac{1}{\left\Vert \mathbf{x}\right\Vert ^{4}}%
\left( 1-\tfrac{1+\alpha }{2\beta }\right) \mathbf{ad}_{\mathbf{X}}^{4} \equationdot
\label{dexpInvSE3ad2}
\end{aligned}%
\end{equation}
The singularity in \reqn{dexpInvSE3ad2} is removable by exploiting the limit $\lim_{\lVert \vect x \rVert \to \vect 0} \mathbf{dexp}_{\mathbf{X}}^{-1} = \mathbf{I}-\frac{1}{2}\mathbf{ad}_{\mathbf{X}}$.

\subsection{Differential of the dexp mapping}
The directional derivative of the matrix $\mathbf{dexp}^{-1}$ for $SE(3)$ is
\begin{align}
(\mathrm{D}_{\mathbf{X}}\mathbf{dexp}^{-1})%
\hspace{-0.5ex}%
\left( \mathbf{U}\right) = &\notag \\
\left( 
\begin{array}{cc}
(\mathrm{D}_{\mathbf{x}}\mathbf{dexp}^{-1})%
\hspace{-0.5ex}%
\left( \mathbf{u}\right) & \mathbf{0} \\ 
(\mathrm{D}_{\mathbf{X}}\mathbf{Ddexp}^{-1})%
\hspace{-0.5ex}%
\left( \mathbf{U}\right) & (\mathrm{D}_{\mathbf{x}}\mathbf{dexp}^{-1})%
\hspace{-0.5ex}%
\left( \mathbf{u}\right)%
\end{array}%
\right)  \label{diffDexpInvSE3}
\end{align}%
where $\mathbf{U}=\left( \mathbf{u},\mathbf{v}\right) $, and the directional
derivative of matrix $\mathbf{Ddexp}^{-1}$ possesses the explicit closed-form%
\begin{align}
(\mathrm{D}_{\mathbf{X}}\mathbf{Ddexp}^{-1})%
\hspace{-0.5ex}%
\left( \mathbf{U}\right) = -\tfrac{1}{2}\tilde{\mathbf{v}} \notag \\
+\tfrac{1-\gamma 
}{\left\Vert \mathbf{x}\right\Vert ^{2}}\left( \tilde{\mathbf{x}}\tilde{%
\mathbf{v}}+\tilde{\mathbf{v}}\tilde{\mathbf{x}}+\tilde{\mathbf{y}}\tilde{%
\mathbf{u}}+\tilde{\mathbf{u}}\tilde{\mathbf{y}}+\tfrac{1}{4}(\mathbf{x}^{T}%
\mathbf{u})(\tilde{\mathbf{x}}\tilde{\mathbf{y}}+\tilde{\mathbf{y}}\tilde{%
\mathbf{x}})\right)  \label{diff2dexpInvSO31} \notag \\
-\tfrac{1}{\left\Vert \mathbf{x}\right\Vert ^{4}}%
\Big%
(\left( 1-\gamma \right) (\mathbf{x}^{T}\mathbf{u})(2+\gamma )(\tilde{%
\mathbf{x}}\tilde{\mathbf{y}}+\tilde{\mathbf{y}}\tilde{\mathbf{x}})  \notag
\\
+(2-\gamma -\tfrac{1}{\beta })(\mathbf{x}^{T}\mathbf{y})(\tilde{\mathbf{x}}%
\tilde{\mathbf{u}}+\tilde{\mathbf{u}}\tilde{\mathbf{x}})+(\mathbf{x}^{T}%
\mathbf{v}+\mathbf{y}^{T}\mathbf{u}) \notag \\
(2+\tfrac{1}{4}(\mathbf{x}^{T}\mathbf{y}%
)(\mathbf{x}^{T}\mathbf{u})-\gamma -\tfrac{1}{\beta })\tilde{\mathbf{x}}^{2}%
\Big%
)  \notag \\
+\tfrac{1}{\left\Vert \mathbf{x}\right\Vert ^{6}}(\mathbf{x}^{T}\mathbf{y}%
)(\mathbf{x}^{T}\mathbf{u})\left( 8-3\gamma -\gamma ^{2}-\tfrac{2}{\beta ^{2}%
}\left( \alpha +\beta \right) \right) \tilde{\mathbf{x}}^{2} \notag \equationdot 
\end{align}
The above relation can be implemented to cope with $\lVert \vect x \rVert = \vect 0$~\cite{muller2021review}. An equivalent expression for the directional derivative of the matrix of the left-trivialized differential, i.e. with negative argument $\vect X$, was derived in Appendix A.1 of~\cite{2015_valentin_thesis}.
Evaluating \reqn{diffDexpInvSE3} additionally requires the directional derivative of $\mathbf{dexp}^{-1}$ for $SO(3)$ which is given by
\begin{align}
 (\mathrm{D}_{\mathbf{x}}\mathbf{dexp}^{-1})(\mathbf{y}) = &
 -\frac{1}{2} \tilde{\mathbf{y}} + 
 \frac{1}{{\lVert \mathbf{x}\rVert}^{2}}\left(1-\gamma\right)(\tilde{\vect x}\tilde{\vect y} + \tilde{\vect y}\tilde{\vect x}) \notag \\
& + \frac{\vect x^T \vect y}{{\lVert \mathbf{x}\rVert}^{4}}\left(\frac{1}{\beta}+\gamma-2\right)\tilde{\vect x}^2 \equationdot
\end{align}


\section{Geometric Linearization of EOM of Floating-Base Systems}
\label{sec_geom_lin_eom}
This section presents the equation of motion of a free-floating single rigid body in $SE(3)$, its state space form, and the linearization for inclusion in finite horizon LQR controllers. Note that we consider a fully actuated free-floating rigid body. However, without any loss of generality, the derivation presented below can be extended to underactuated free-floating systems by including an actuator selection matrix.

\subsection{EOM of free-floating single rigid body}
Let us consider a simple case of free-floating rigid body with its body fixed reference (BFR) frame located at the center of mass (COM). 
The EOM of such a rigid body in $SE(3)$ is given by
\begin{equation}
 \vect W = \mat M \dot{\vect V} + \mathbf{ad}_{\vect V}^T \mat M \vect V 
 \label{eqn_eom_rigid_body}
\end{equation}
where $\mat W \in se^*(3)$ is the net wrench acting on the body, $\vect V \in se(3)$ and $\dot{\vect V} \in \mathbb{R}^6$ represent the twist and acceleration of the moving body respectively - all in body fixed representation. $\mat M \in \mathcal{P}(6)$ denotes the $6 \times 6$ symmetric and positive-definite mass-inertia matrix of the body with the following form:
\begin{equation}
 \mat M = 
\left[
  \begin{matrix}
  \vect I_b & \vect 0 \\
  \vect 0   & m \mat I \\
  \end{matrix}
\right] 
\end{equation}
where $\mat I_b \in \mathcal{P}(3)$ is its rotational inertia and $m \in \mathbb{R}^+$ is the mass of the moving body. The expression for forward dynamics can be obtained by rearranging \reqn{eqn_eom_rigid_body} as
\begin{equation}
 \dot{\vect V} = \mat M^{-1} \left(\vect W - \mathbf{ad}_{\vect V}^T \mat M \vect V \right) 
 \label{eqn_fdyn}
\end{equation}
which is a $2^{\text{nd}}$ order ordinary differential equation (ODE). 

\subsection{Dynamics in state-space representation}
Let $\vect S = (\bm \eta, \bm \xi)^T \in \mathbb{R}^6$ denote the canonical screw coordinates of $SE(3)$. Together with $\vect V$, one could denote the state of the rigid body as $\vect x = (\vect S, \vect V)^T \in \mathbb{R}^{12}$. Using \reqn{eqn_fdyn} and $\vect V = \mathbf{dexp}_{-\vect S} \dot{\vect S} $ (substitute $\vect X = \vect S$ in \reqn{RecExp}), its first order time derivative $\dot{\vect x} \in \mathbb{R}^{12}$ is given by:
\begin{equation}
 \dot{\vect x} =
\left[
  \begin{matrix}
  \dot{\vect S}  \\
  \dot{\vect V}    \\
  \end{matrix}
\right]  =
\left[
  \begin{matrix}
  \mathbf{dexp}_{-\vect S}^{-1} \vect V     \\
  \mat M^{-1} \left(\vect W - \mathbf{ad}_{\vect V}^T \mat M \vect V \right)  \\  
  \end{matrix} 
\right] =
\vect f (\vect x, \vect W)
\label{eqn_dyn_1st_order_ode}
\end{equation}
which captures the dynamics of the free-floating rigid body in the form of a $1^{\text{st}}$ order ODE. Note that any other choice of coordinates of $SE(3)$ (Study parameters~\cite{2007_Husty}, dual quaternions~\cite{blaschke1958anwendung}) would require resolution of additional algebraic constraints typically leading to a differential-algebraic equation (DAE).

\subsection{Linearization of the state-space dynamics}
Considering the Taylor series expansion of \reqn{eqn_dyn_1st_order_ode}, the system dynamics can be linearized and written in the following state-space form:
\begin{equation}
 \dot{\vect x} = \mat A \vect x + \mat B \vect W
\end{equation}
where $\mat A = \frac{\partial \vect f}{\partial \vect x} \in \mathbb{R}^{12 \times 12}$ and $\mat B = \frac{\partial \vect f}{\partial \vect W} \in \mathbb{R}^{12 \times 6}$ are the partial derivatives of the dynamics $\vect f(\vect x, \vect W)$ with respect to the state vector $\vect x$ and wrench acting on the body $\vect W$ respectively. The expression for the matrix $\mat A$ is given by
\begin{equation}
 \mat A = 
\left[
  \begin{matrix}
  \frac{\partial \left(\mathbf{dexp}^{-1}_{-\vect S} \vect V \right)}{\partial \vect S} & \mathbf{dexp}^{-1}_{-\vect S} \\
  \mat 0_{6 \times 6}   & \mat M^{-1} \frac{\partial\left( \mathbf{ad}_{\vect V}^T \mat M \vect V \right)}{\partial \vect V} \\
  \end{matrix}
\right] \equationdot
\label{eqn_mat_A}
\end{equation}
Here the top-left block matrix $\frac{\partial \left(\mathbf{dexp}^{-1}_{-\vect S} \vect V \right)}{\partial \vect S}$ requires the multiplication of first order partial differential of the inverse of $\mathbf{dexp}$ mapping (which is a tensor) by $\vect V$. In order to avoid the computation of this tensor quantity explicitly, one can exploit the directional derivative expression in \reqn{diffDexpInvSE3} with the basis vectors $\vect e_i$ taken from the $i^\text{th}$ column of a $6 \times 6$ identity matrix $\mat I_{6 \times 6}$ (for $i \in \{1,2,\ldots,6\}$) as
\begin{align*}
\frac{\partial \left(\mathbf{dexp}^{-1}_{-\vect S} \vect V \right)}{\partial \vect S} = \notag \\
\left[
  \begin{matrix}
(\mathrm{D}_{\mathbf{-S}}\mathbf{dexp}^{-1})\left( \mathbf{e}_1\right) \vect V &
\ldots &
(\mathrm{D}_{\mathbf{-S}}\mathbf{dexp}^{-1})\left( \mathbf{e}_6\right) \vect V
  \end{matrix}
\right]_{6 \times 6} \equationdot
\end{align*}
The top-right block matrix can be evaluated with \reqn{dexpInvSE3ad2}. The bottom-right block matrix $\mat M^{-1} \frac{\partial\left( \mathbf{ad}_{\vect V}^T \mat M \vect V \right)}{\partial \vect V}$ requires the use of tensorial quantities but can be computed easily by multiplying $\mat M^{-1}$ with the formula for
\begin{equation*}
\begin{aligned}
\frac{\partial\left( \mathbf{ad}_{\vect V}^T \mat M \vect V \right)}{\partial \vect V} & = 
\mathbf{ad}_{\vect V}^T \mat M + \\
& \left ( \mathbf{ad}_{\vect e_1}^T \mat M \vect V \quad \mathbf{ad}_{\vect e_2}^T \mat M \vect V \quad \ldots \quad \mathbf{ad}_{\vect e_6}^T \mat M \vect V \right) \equationdot
\end{aligned} 
\end{equation*}
In the second matrix summand of the above formula, each matrix column $\mathbf{ad}_{\vect e_i}^T \mat M \vect V$ is evaluated with the basis vectors $\vect e_i$ as before. The expression of the matrix $\mat B$ is simply given by
\begin{equation}
 \mat B = 
\left[
  \begin{matrix}
  \mat 0  \\
  \mat M^{-1}    \\
  \end{matrix}
\right] \equationdot
\label{eqn_mat_B}
\end{equation}

\section{Trajectory stabilization in $SE(3)$}
\label{sec_tvlqr_se3}
Assume that pre-computed optimal state $\vect x_0 (t) = (\vect S_0(t), \vect V_0(t))^T$ and input $(\vect W_0(t))$ trajectories for the floating base system are described in terms of canonical coordinates (screw coordinates) on $SE(3)$. The error in a local coordinate system relative to the nominal trajectory can be defined as:
\begin{equation*}
 \bar{\vect x}(t) = \vect x(t) - \vect x_0(t) \in \mathbb{R}^{12} \equationcomma \quad
 \bar{\vect W}(t) = \vect W(t) - \vect W_0(t) \in \mathbb{R}^{6}
\end{equation*}
The time derivative of the state error trajectory is
\begin{equation*}
 \dot{\bar{\vect x}}(t) = \dot{\vect x}(t) - \dot{\vect x}_0(t) = 
 \vect f(\vect x(t), \vect W(t)) - \vect f(\vect x_0(t), \vect W_0(t))
\end{equation*}
which can be approximated by $1^\text{st}$ order Taylor series approximation as
\begin{equation*}
\begin{aligned}
 \dot{\bar{\vect x}}(t) & \approx \vect f(\vect x_0(t), \vect W_0(t)) +  
 \frac{\partial \vect f (\vect x_0(t), \vect W_0(t))}{\partial \vect x} (\vect x(t) - \vect x_0(t)) \\
& + \frac{\partial \vect f (\vect x_0(t), \vect W_0(t))}{\partial \vect W} (\vect W(t) - \vect W_0(t)) -
\vect f(\vect x_0(t), \vect W_0(t)) \\
& = \mat A(t) \bar{\vect x}(t) + \mat B(t) \bar{\vect u}(t)
\end{aligned}
\end{equation*}
where $\mat A(t)$ and $\mat B(t)$ can be evaluated with \reqn{eqn_mat_A} and \reqn{eqn_mat_B} respectively.
Note that the linearization is time varying and is evaluated along the nominal trajectory. 

Considering a quadratic form on the trajectory following cost
\begin{equation*}
J = \bar{\vect x}^T(t) \mat Q_f \bar{\vect x}(t) + \int_0^{t_f} \left( \bar{\vect x}^T(t) \mat Q \bar{\vect x}(t) + \bar{\vect W}(t) \mat R \bar{\vect W}(t) \right) dt
\end{equation*}
with $\mat Q \succeq 0, \mat Q_f \succeq 0$ and $\mat R \succ 0$, the optimal tracking controller can be derived by solving differential Riccati equation
\begin{align}
-\dot{\mat S}(t)
      =& {\mat S}(t) {\mat A}(t) + {\mat A}^T(t) {\mat S}(t) \notag \\ 
     & - {\mat S}(t) \mat B(t) {\mat R}^{-1}
      \mat B^T(t) {\mat S}(t) + {\mat Q}, 
      \label{eqn_diff_ricatti}
\end{align}
with the terminal condition
\begin{equation}
 \mat S(t_f) = \mat Q_f \equationdot
\end{equation}
The resulting optimal controller takes the linear form $\bar{\vect W}^*(t) = -\mat K(t) \bar{\vect x}(t)$, or:
\begin{equation}
 \vect W^*(t) = \vect W_0(t) - \mat K(t) (\dot{\vect x}(t) - \dot{\vect x}_0(t))
\end{equation}
and the corresponding optimal cost-to-go function is given by
\begin{equation}
 J^*(\vect x(t)) = \bar{\vect x}^T(t) \mat S(t) \bar{\vect x}(t) \equationdot
\end{equation}
Note that it is crucial to respect the symmetric and positive semi-definite structure of the $\mat S(t)$ matrix in order to avoid numerical errors in solving \reqn{eqn_diff_ricatti}. This can be done by using the square-root factorization of the matrix $\mat S(t) = \mat P^T(t) \mat P(t)$ as discussed in~\cite{underactuated}.
An alternative approach to ensure this is to use symplectic integrators as shown in~\cite{dieci1994positive}.

\section{Conclusion and Outlook}
\label{sec_conc_outlook}
This paper presents a novel geometric linearization of the equations of motion of a free-floating rigid body in $SE(3)$ in terms of its canonical screw coordinates. This linearization exploits the differential of the exponential map and its directional derivative in order to compute the involved partial derivatives. Subsequently, we use this linearization to develop a finite horizon LQR controller which can be used to locally stabilize any pre-computed optimal trajectory. The next step includes numerical validation of the approach presented in this paper. In the near future,  we also plan to extend our work to other similar control approaches, such as iterative LQR (iLQR) and differential dynamic programming (DDP).




\bibliography{references}

\end{document}